\documentclass[runningheads,a4paper]{llncs}

\usepackage{amssymb}
\usepackage{graphicx}

\usepackage{url}
\urldef{\mailsa}\path|{alfred.hofmann, ursula.barth, ingrid.haas, frank.holzwarth,|
\urldef{\mailsb}\path|anna.kramer, leonie.kunz, christine.reiss, nicole.sator,|
\urldef{\mailsc}\path|erika.siebert-cole, peter.strasser, lncs}@springer.com|

\usepackage{algorithm}
\usepackage{algorithmic}

\usepackage{hyperref}

\usepackage{xspace}
\usepackage{amsfonts,amstext,amsmath,amssymb}
\usepackage{multirow}
\usepackage{array,url}
\usepackage{color}

\newcommand{\fb}{{\sc Freebase}\xspace}

\newcommand{\wk}{{\sc WikiAnswers}\xspace}
\newcommand{\wq}{{\sc WebQuestions}\xspace}

\newcommand{\cw}{{\sc ClueWeb}\xspace}

\newcommand{\us}{Our approach}

\newcommand{\ent}[1]{{\small {\tt #1}}}

\begin{document}

\mainmatter  

\title{Question Answering with Subgraph Embeddings }

\titlerunning{Question Answering with Subgraph Embeddings}

%
%
\author{Antoine Bordes
\and Jason Weston \and Sumit Chopra}
%

\institute{Facebook, 770 Broadway, New York, NY, USA \\
\path|{{abordes,jase,spchopra}@fb.com}|
}

%
%

\toctitle{Lecture Notes in Computer Science}
\tocauthor{Authors' Instructions}
\maketitle

\begin{abstract}
This paper presents a system which learns to answer questions on a
broad range of topics from a knowledge base using few hand-crafted features.
%
Our model learns low-dimensional embeddings of words and
knowledge base constituents; these representations
are used to score natural language questions against candidate answers.
%
Training our system using pairs of questions and structured
representations of their answers, and pairs of
question paraphrases, yields competitive results on a
recent benchmark of the literature.
\end{abstract}

\section{Introduction}

Teaching machines how to automatically answer questions asked in natural
language on any topic or in any domain has always been a long standing
goal in Artificial Intelligence.
With the rise of large scale structured knowledge bases (KBs), this
problem, known as open-domain question answering (or open QA), 
boils down to being able to
query efficiently such databases with natural language.
These KBs, such as \fb \cite{bollacker2008freebase} 
encompass huge ever growing amounts of information
and
ease open QA by organizing a great variety of answers in a structured
format.
However, the scale and the difficulty for machines to
interpret natural language still makes this task a challenging problem.

The state-of-the-art techniques in open QA can be classified 
into two main classes, namely, information retrieval based and 
semantic parsing based. 
Information retrieval systems first retrieve a broad set
of candidate answers by querying the search API of KBs with a
transformation of the question into a valid query and then use
fine-grained detection heuristics to identify the exact answer
\cite{kolomiyets2011survey,unger2012template,yao2014information}.
On the other hand, semantic parsing methods focus on the correct
interpretation of the meaning of a question by a semantic parsing
system. A correct interpretation converts a question
into the exact database query that returns the correct answer.
Interestingly, recent works
\cite{berant-EtAl:2013:EMNLP,kwiatkowski-EtAl:2013:EMNLP,berant2014semantic,fader2014open}
have shown that such systems can be efficiently trained under
indirect and imperfect supervision and hence scale to large-scale
regimes, while bypassing most of the annotation costs.

Yet, even if both kinds of system have shown the ability to handle
large-scale KBs, they still require experts to hand-craft lexicons,
grammars, and KB schema to be effective. 
This non-negligible human
intervention might not be generic enough to conveniently scale up to
new databases with other schema, broader vocabularies or 
languages other than English. 
In contrast, \cite{paralex} proposed a framework for open QA requiring
almost no human annotation.
Despite being an interesting approach, this method is outperformed by other
competing methods.
%
\cite{bordes2014open}  introduced an embedding model, which
learns low-dimensional vector representations of words and 
symbols
(such as KBs constituents) and can be trained
with even less supervision than the system of \cite{paralex} while
being able to achieve better prediction performance.
%
However, this approach is only compared with \cite{paralex} which
operates in a simplified setting and has not been applied in 
more realistic conditions nor evaluated against the best performing
methods.

In this paper, we improve the model of \cite{bordes2014open} by
providing the ability to answer more complicated questions. 
s%
%
The main contributions of the paper are: (1) a more sophisticated 
inference procedure that is both efficient and can consider longer paths (\cite{bordes2014open} considered only answers directly connected to the question in the graph); and (2) a richer
representation of the answers which encodes the question-answer path and surrounding
subgraph of the KB. 
Our approach is competitive with the current state-of-the-art 
on the recent benchmark \wq
\cite{berant-EtAl:2013:EMNLP} without using any lexicon, rules or additional 
system for part-of-speech tagging, syntactic or dependency parsing
during training as most other systems do.


\section{Task Definition} \label{data}

Our main motivation is to provide a system for open QA able to be
trained as long as it has access to: (1) a training set of questions
paired with answers and (2) a KB providing a structure among answers.
We suppose that all potential answers are entities in the KB and that
questions are sequences of words that include one identified KB
entity. When this entity is not given, plain string matching is used to
perform entity resolution. Smarter methods could be used but
this is not our focus.

We use \wq \cite{berant-EtAl:2013:EMNLP} as our evaluation
bemchmark.
Since it contains few training samples, it is impossible to learn on
it alone, and this section describes the various data sources that
were used for training. These are similar to those used in
\cite{berant2014semantic}.

\paragraph{WebQuestions}
This dataset is built using \fb as the KB and contains 5,810
question-answer pairs.  It was created by crawling questions through
the Google Suggest API, and then obtaining answers using Amazon
Mechanical Turk.  We used the original split (3,778 examples for
training and 2,032 for testing), and isolated 1k questions from the
training set for validation.
\wq is built on \fb since all answers are defined as \fb entities.
In each question, we identified one \fb entity using string matching between words
of the question and entity names in \fb.
When the same string matches multiple entities, only the entity
appearing in most triples, i.e. the most popular in \fb, was kept.
%
%
%
%
Example questions (answers) in the dataset include 
{\em ``Where did Edgar  Allan Poe died?''} (\ent{baltimore}) or  {\em ``What degrees did
  Barack Obama get?''} (\ent{bachelor\_of\_arts}, \ent{juris\_doctor}).


\paragraph{Freebase}
\fb \cite{bollacker2008freebase} is a huge and freely available
database of general facts; data is organized as triplets
(\ent{subject}, \ent{type1.type2.predicate}, \ent{object}), where two
entities \ent{subject} and \ent{object} (identified by \ent{mids}) are
connected by the relation type \ent{type1.type2.predicate}.
%
%
We used a subset, created by only keeping
triples where one of the entities was appearing in either the \wq
training/validation set or in  \cw extractions.
We also removed all entities appearing less than 5 times and finally
obtained a \fb set containing 14M triples made of 2.2M entities and 7k
relation types.\footnote{\wq contains $\sim$2k entities, hence
  restricting \fb to 2.2M entities does not ease the task for us.}
%
%
Since the format of triples does not correspond to any structure one could find in
language, we decided to transform them into automatically generated
questions.  Hence, all triples were converted into questions ``What is
the \ent{predicate} of the \ent{type2} \ent{subject}?'' (using the
\ent{mid} of the subject) with the answer being \ent{object}.  
An example is {\em ``What is the \ent{nationality} of
the \ent{person} \ent{barack\_obama}?''} (\ent{united\_states}). 
More examples and details are given in a longer version of this paper  \cite{DBLP:journals/corr/BordesCW14}.

\begin{table}[t!]
\begin{center}
\begin{small}
\begin{tabular}{|ll|r|}
\hline 
\wq & -- Train. ex. & 2,778\\
& -- Valid. ex. & 1,000\\
& -- Test. ex. & 2,032\\
\fb & -- Train. ex. & 14,790,259 \\
\cw & -- Train. ex. & 2,169,033 \\
\wk & -- Train. quest. & 2,423,185 \\
 & -- Parap. clust. & 349,957 \\
\hline 
Dictionary & -- Words & 1,526,768\\
& -- Entities &  2,154,345 \\
& -- Rel. types & 7,210 \\ 
\hline 
\end{tabular}
\end{small}
\caption{\label{tab:stats} Statistics of data sets used in the paper.}
\end{center}
\end{table}

\paragraph{ClueWeb Extractions}

\fb data allows to train our model on 14M questions but these have a
fixed lexicon and vocabulary, which is not realistic.
Following  \cite{berant-EtAl:2013:EMNLP}, we also created questions
using \cw extractions provided by \cite{lin2012entity}.
%
%
Using string matching, we ended up with 2M extractions structured as
(\ent{subject}, ``text string'', \ent{object}) with both \ent{subject} and
\ent{object} linked to \fb.
We also converted these triples into questions by using simple patterns and
\fb types. 
An example of generated question is {\em ``Where
\ent{barack\_obama} was allegedly bear in?''} (\ent{hawaii}).

\paragraph{Paraphrases}

The automatically generated questions that are useful to connect \fb
triples and natural language, do not provide a satisfactory
modeling of natural language because of their semi-automatic wording
and rigid syntax.
To overcome this issue, we follow \cite{paralex} and supplement our training data with an
indirect supervision signal made of pairs of question paraphrases
collected from the \wk website.
On \wk, users can tag pairs of questions as rephrasings of each other:
\cite{paralex} harvested a set of 2M distinct questions from
\wk, which were grouped into 350k paraphrase clusters.
%

\begin{table*}[t!]
\begin{center}
\resizebox{\linewidth}{!}{
\begin{tabular}{|l|l|}
\hline 
& what is the judicial capital of the in state \ent{sikkim} ? -- \ent{gangtok}\\
& (\ent{sikkim}, \ent{location.in\_state.judicial\_capital},  \ent{gangtok}) \\ 
\cline{2-2}
&   who influenced the influence node \ent{yves\_saint\_laurent} ? -- \ent{helmut\_newton}\\
&  (\ent{yves\_saint\_laurent}, \ent{influence.influence\_node.influenced}, \ent{helmut\_newton})  \\
\cline{2-2}
\fb &   who is born in the location \ent{brighouse} ? -- \ent{edward\_barber} \\
{\it generated questions} & (\ent{brighouse}, \ent{location.location.people\_born\_here}, \ent{edward\_barber})  \\
\cline{2-2}
{\it and associated triples} & who is the producer of the recording
\ent{rhapsody\_in\_b\_minor,\_op.\_79,\_no.\_1} ? -- \ent{glenn\_gould}\\
& (\ent{rhapsody\_in\_b\_minor,\_op.\_79,\_no.\_1},
\ent{music.recording.producer}, \ent{glenn\_gould})  \\
\cline{2-2}
&   what are the symptoms of the disease \ent{sepsis} ? -- \ent{skin\_discoloration}\\
& (\ent{sepsis}, \ent{medicine.disease.symptoms}, \ent{skin\_discoloration}) \\
\hline 
\hline 
& what is \ent{cher}'s son's name ? -- \ent{elijah\_blue\_allman}\\
& (\ent{cher}, \ent{people.person.children},
\ent{elijah\_blue\_allman})\\
\cline{2-2}
& what are dollars called in \ent{spain} ? -- \ent{peseta} \\
& (\ent{spain}, \ent{location.country.currency\_formerly\_used}, \ent{peseta}) \\        
\cline{2-2}
\wq & what is \ent{henry\_clay} known for ? -- \ent{lawyer} \\
{\it training questions} & (\ent{henry\_clay}, \ent{people.person.profession}, \ent{lawyer}) \\
\cline{2-2}
{\it and associated paths} & who is the president of the
\ent{european\_union} 2011 ? -- \ent{jerzy\_buzek}\\
& (\ent{european\_union}, \ent{government.governmental\_jurisdiction.governing\_officials}\\
&\ent{government.government\_position\_held.office\_holder},  \ent{jerzy\_buzek}) \\ 
\cline{2-2}
& what 6 states border \ent{south\_dakota} ? -- \ent{iowa}\\
& (\ent{south\_dakota}, \ent{location.location.contains}
\ent{location.location.partially\_containedby},  \ent{iowa}) \\
\hline
\hline
& what does \ent{acetazolamide} be an inhibitor of ? -- \ent{carbonic\_anhydrase}\\
& (\ent{acetazolamide}, \ent{medicine.drug\_ingredient.active\_moiety\_of\_drug}, \ent{carbonic\_anhydrase})\\
\cline{2-2}
& which place is a district in \ent{andhra\_pradesh} ? -- \ent{adilabad}\\
& (\ent{andhra\_pradesh}, \ent{location.location.contains}, \ent{adilabad}) \\
\cline{2-2}
\cw & what is a regional airline based in \ent{dublin} ? -- \ent{aer\_arann} \\
{\it generated questions} & (\ent{dublin}, \ent{location.location.nearby\_airports} \ent{aviation.airport.focus\_city\_for},  \ent{aer\_arann})\\
\cline{2-2}
{\it and associated paths} & what mean fire in \ent{sanskrit} ? -- \ent{agni}\\
& (\ent{sanskrit}, \ent{unknown\_relation}, \ent{agni}) \\
\cline{2-2}
& where does \ent{american\_legion} proceed to ? -- \ent{san\_francisco}\\  
& (\ent{american\_legion}, \ent{22-rdf-syntax-ns\#type} \ent{type.type.instance}, \ent{san\_francisco})  \\
\hline 
\hline 
& what are two reason to get a 404 ? \\
& what is error 404 ? \\
& you receive a 404 - unable to find error message ?\\
& how do you correct error 404 ?\\
\cline{2-2}
\wk & what is the term for a teacher of islamic law ?\\
{\it clusters of } & what is the islamic religious teacher called ?\\
{\it quest. paraphrases } & what is the name of the religious book islam use ?\\
& who is chief of islamic religious authority ?\\
\cline{2-2}
& what  country is bueno aire in ?\\
& what countrie is buenos aires in ?\\
& what country is buenas aire in ?\\
&  what country is bueno are in ?\\
\hline 
\end{tabular}
}
\caption{\label{tab:ex} Examples of questions, answer paths and paraphrases  used in this
paper.}
\end{center}
\end{table*}

\section{Embedding Questions and Answers} \label{model} 

Inspired
by~\cite{bordes2014open}, our model works by learning low-dimensional
vector embeddings of words appearing in questions and of
entities and relation types of \fb, so that representations of 
questions and of their corresponding answers are close to each other
in the joint embedding space. 
Let $q$ denote a question and $a$ a candidate answer.  
Learning embeddings is achieved by learning a scoring function $S(q, a)$,  
so that $S$ generates a high score if $a$ is the correct 
answer to the question $q$, and a low score otherwise. Note that both $q$ and $a$ 
are represented as a combination of the embeddings of their individual words and/or 
symbols; hence, learning $S$ essentially involves learning these 
embeddings. In our model, the form of the scoring function is: 
\begin{equation} \label{rank-eq}
S(q, a) = f(q)^{\top}g(a). 
\end{equation}
Let $\bf{W}$ be a matrix of $\mathbb{R}^{k \times N}$, 
where $k$ is the dimension of the embedding space which is fixed a-priori,
and $N$ is the dictionary of embeddings to be learned.
Let $N_{W}$ denote the total number of words
and $N_{S}$ the total number of entities and relation types.
With $N = N_{W} + N_{S}$, the
 $i$-th column of ${\bf W}$ is the embedding of the $i$-th 
element (word, entity or relation type) in the dictionary. 
The function $f(.)$, which maps the questions into the embedding 
space $\mathbb{R}^k$ is defined as $f(q) = {\bf W}\phi(q)$, where 
$\phi(q) \in \mathbb{N}^N$, is a sparse vector indicating the 
number of times each word appears in the question $q$ (usually 0 or 1). 
Likewise the function $g(.)$ which maps the answer 
into the same embedding space $\mathbb{R}^k$ as the questions, 
is given by $g(a) = {\bf W}\psi(a)$. 
Here $\psi(a) \in \mathbb{N}^N$ is a sparse vector representation 
of the answer $a$, which we now detail. 

\begin{figure*}
\begin{center}
\includegraphics[width=12cm]{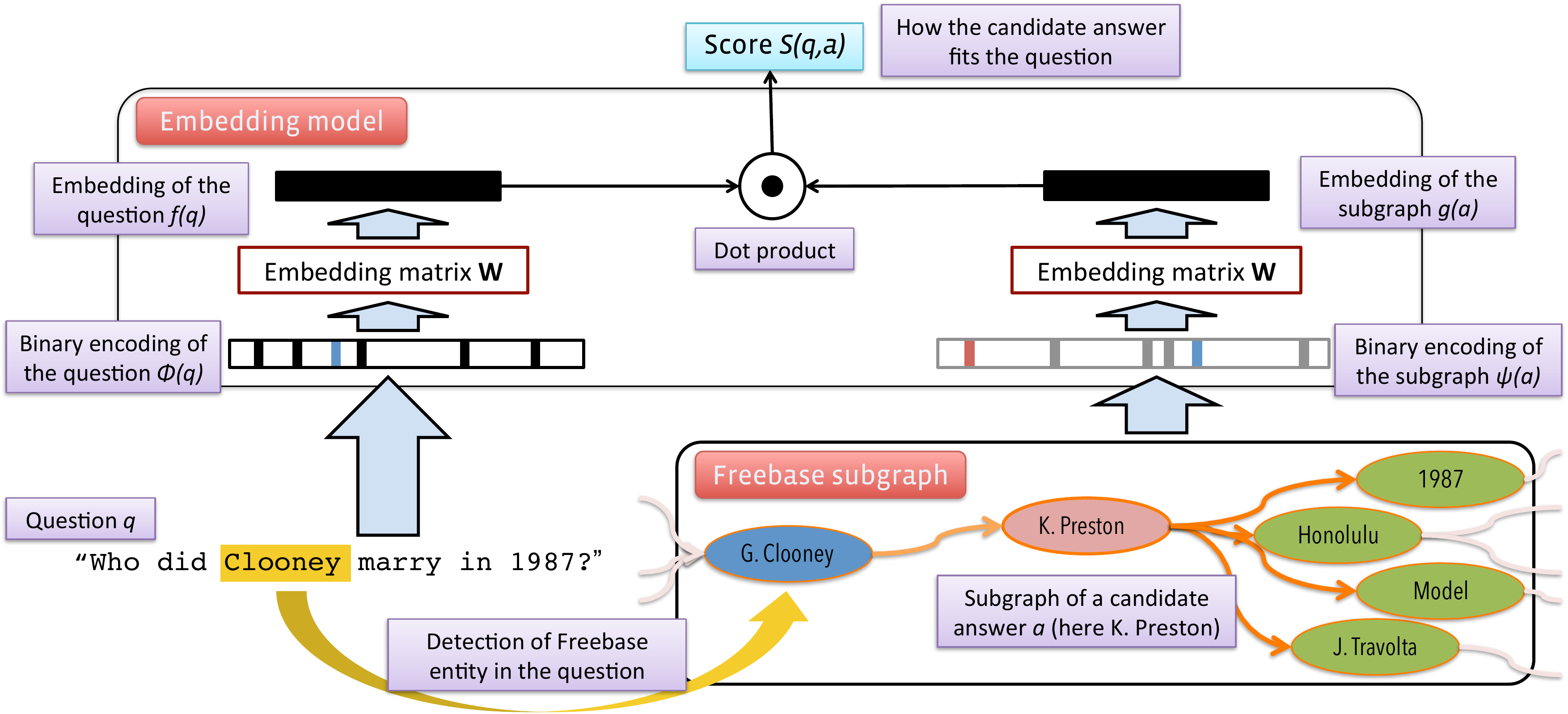}
\caption{\label{fig:subgraph} Illustration of the subgraph embedding
  model scoring a candidate answer: (i) locate entity in the question;
  (ii) compute path from entity to answer; (iii) represent answer as
  path plus all connected entities to the answer (the subgraph); (iv)
  embed both the question and the answer subgraph separately using the
  learnt embedding vectors, and score the match via their dot
  product.}
\end{center}
\end{figure*}

\subsection{Representing Candidate Answers} \label{ans-features}

We now describe possible feature representations for a single
candidate answer. (When there are multiple correct answers, we average these 
representations, see Section \ref{multiple-ans}.)
We consider three different types of representation,
corresponding to different subgraphs of \fb around
it.
%
%
\begin{itemize}
\item[(i)] {\it Single Entity}. The answer is represented 
as a single entity from \fb:
$\psi(a)$ is a 1-of-$N_S$ coded vector with 1 corresponding to the 
entity of the answer, and 0 elsewhere. 
\item[(ii)] {\it Path Representation}. The answer is represented as a
  path from the entity mentioned in the question to the answer
  entity.  In our experiments, we considered 1- or 2-hops paths
  (i.e. with either 1 or 2 edges to traverse):
(\ent{barack\_obama},
\ent{people.}\ent{person.}\ent{place\_of\_birth},~\ent{honolulu}) is a 1-hop path
and  (\ent{barack\_obama},
\ent{people.person.place\_of\_birth},~\ent{location.} \ent{location.containedby}, \ent{hawaii}) a 2-hops path.
  This results in a $\psi(a)$ which is a 3-of-$N_S$ or 4-of-$N_S$ coded
  vector, expressing the start and end entities of the path and the
  relation types (but not entities) in-between.
\item[(iii)] {\it Subgraph Representation}. We encode both the path
  representation from (ii), and the entire subgraph of entities
  connected to the candidate answer entity. That is, for each entity
  connected to the answer we include both the relation type and the
  entity itself in the representation $\psi(a)$. In order to represent
  the answer path differently to the surrounding subgraph (so the
  model can differentiate them), we double the dictionary size for
  entities, and use one embedding representation if they are in the
  path and another if they are in the subgraph.  Thus we now learn a
  parameter matrix $\mathbb{R}^{k \times N}$ where $N = N_{W} + 2
  N_{S}$ ($N_{S}$ is the total number of entities and relation
  types). If there are $C$ connected entities with $D$ relation types
  to the candidate answer, its representation is a $3+C+D$ or
  $4+C+D$-of-$N_S$ coded vector, depending on the path length.
  %

\end{itemize}

\label{sec:hypo}

Our hypothesis is that including more information about the answer in
its representation will lead to improved results. While it is possible
that all required information could be encoded in the $k$ dimensional
embedding of the single entity {\it (i)}, it is unclear what dimension
$k$ should be to make this possible. For example the embedding
of a country entity encoding all of its citizens seems unrealistic.
Similarly, only having access to the path ignores all the other
information we have about the answer entity, unless it is encoded in
the embeddings of either the entity of the question, the answer or the
relations linking them, which might be quite complicated
as well.  We thus adopt the subgraph approach.
Figure~\ref{fig:subgraph} illustrates our model.


\subsection{Training and Loss Function}
As in~\cite{wsabie}, we train our model using a margin-based
ranking loss function.  Let $\mathcal{D} = \{(q_i, a_i): i = 1,
\ldots, |\mathcal{D}|\}$ be the training set of 
questions $q_i$ paired with their correct answer $a_i$. The loss
function we minimize is
\begin{equation} \label{eq:loss}
\sum_{i=1}^{|\mathcal{D}|} \sum_{\tiny{\bar{a} \in \bar{\mathcal{A}}(a_i)}}
\max\{0, m - S(q_i,a_i) + S(q_i,\bar{a})\}, 
\end{equation}
where $m$ is the margin (fixed to $0.1$).  Minimizing
Eq. (\ref{eq:loss}) learns the embedding matrix ${\bf W}$ so that the score
of a question paired with a correct answer is greater than with any
incorrect answer $\bar{a}$ by at least $m$. 
$\bar{a}$ is sampled from a set of incorrect candidates $\bar{\mathcal{A}}$.  
This is achieved by sampling 50\% of the
time from the set of entities connected to the entity of the question
(i.e. other candidate paths), and by replacing the
answer entity by a random one otherwise.
Optimization is accomplished using stochastic gradient descent,
multi-threaded with Hogwild!  \cite{recht2011hogwild}, with the 
constraint that the columns $w_i$ of ${\bf W}$ remain within the unit-ball, 
i.e., $\forall_i, ||w_i||_2 \le 1$.

\subsection{Multitask Training of Embeddings}
Since a large number of questions in our training datasets are
synthetically generated, they do not adequately
cover the range of syntax used in natural language.
Hence, we also multi-task the training of our model 
with the task of paraphrase prediction. 
We do so by alternating the training of $S$ with that of a scoring
function $S_{prp}(q_1, q_2) = f(q_1)^\top f(q_2)$, which uses the same
embedding matrix ${\bf W}$ and makes the embeddings of a pair of questions
$(q_1, q_2)$ similar to each other if they are paraphrases (i.e. if
they belong to the same paraphrase cluster), and make them different
otherwise. 
Training $S_{prp}$ is similar to that of $S$ except that negative samples are 
obtained by sampling a question from another paraphrase cluster.

We also multitask the training of the embeddings with the
mapping of the \ent{mids} of \fb entities to the actual words of their
names, so that the model learns that the embedding
of the \ent{mid} of an entity should be similar to the embedding of the
word(s) that compose its name(s).


\subsection{Inference} \label{inference} \label{multiple-ans}
Once ${\bf W}$ is trained, at test time, for a given
question $q$ the model predicts the answer with:
\begin{equation}\label{eq:inf}
\hat{a} = {\mbox{argmax}}_{a' \in \mathcal{A}(q)} S(q,a')
\end{equation}
where $ \mathcal{A}(q)$ is the candidate answer set.
This candidate set could be the whole KB but this has both speed and
potentially precision issues. Instead, we create a candidate set 
$\mathcal{A}(q)$ for each question. 

We recall that each question contains
one identified \fb entity.
$\mathcal{A}(q)$ is first populated with all triples from \fb
involving this entity. This allows to answer simple factual questions
whose answers are directly connected to them (i.e. 1-hop paths). This
strategy is denoted $C_1$.
%

Since a system able to answer only such questions would be limited, we
supplement $\mathcal{A}(q)$ with examples situated in the KB graph at
2-hops from the entity of the question. We do not add all such
quadruplets since this would lead to very large candidate sets. Instead,
we consider the following general approach: given that we are
predicting a path, we can predict its elements in turn using a beam
search, and hence avoid scoring all candidates.
Specifically, our model first ranks relation types using
Eq. (\ref{rank-eq}), i.e.  selects which relation types are the most
likely to be expressed in $q$. We keep the top $10$ types ($10$ was
selected on the validation set) and only add 2-hops candidates to
$\mathcal{A}(q)$ when these relations appear in their path. Scores of
1-hop triples are weighted by $1.5$ since they have one less
element than 2-hops quadruplets.  This strategy, denoted $C_2$, is used by
default.

A prediction $a'$ can commonly actually be a set of candidate answers,
not just one answer, for example for questions like {\em ``Who are David
Beckham's children?''}. This is achieved by considering a prediction to
be all the entities that lie on the same 1-hop or 2-hops path from the
entity found in the question. 
Hence, all answers to the above question are connected to
\ent{david\_beckham} via the same path (\ent{david\_beckham},
\ent{people.person.children}, \ent{*}).
The feature representation of the prediction is then the average over
each candidate entity's features (see Section \ref{ans-features}),
i.e. $\psi_{all}(a') = \frac{1}{|a'|} \sum_{a'_j : a'} \psi(a'_j)$
where $a'_j$ are the individual entities in the overall prediction
$a'$. In the results, we compare to a baseline method that can only
predict single candidates, which understandly performs poorly.

\section{Experiments} 

\begin{table}[t!]
\begin{center}
\begin{tabular}{|@{\,\,}l@{\,}|c|@{}c@{}|c|}
\hline
Method & P@1 & F1 & F1\\
& (\%) & (Berant) & (Yao)\\
\hline
{\bf Baselines} &&&\\
(Berant et al., 2013) \cite{berant-EtAl:2013:EMNLP} & -- & 31.4 & -- \\
(Bordes et al., 2014) \cite{bordes2014open} &  31.3 & 29.7 & 31.8 \\
(Yao and Van Durme, 2014)  \cite{yao2014information} & -- & 33.0 & 42.0\\
(Berant and Liang, 2014) \cite{berant2014semantic} & -- & 39.9 & 43.0\\
\hline
{\bf \us} & & & \\
Subgraph \& $\mathcal{A}(q)=C_2$  & \bf 40.4 & 39.2 & 43.2\\
Ensemble with (Berant \& Liang, 14) & -- &\bf 41.8 &\bf 45.7\\
\hline
{\bf Variants} & & & \\
Without multiple predictions  &\bf  40.4 & 31.3 & 34.2 \\
Subgraph \& $\mathcal{A}(q)=\mbox{All 2-hops}$ & 38.0 & 37.1 & 41.4\\
Subgraph \&  $\mathcal{A}(q)=C_1$ & 34.0 & 32.6 & 35.1\\
Path  \& $\mathcal{A}(q)=C_2$ & 36.2 & 35.3 & 38.5\\
Single Entity  \& $\mathcal{A}(q)=C_1$ & 25.8 & 16.0 & 17.8\\
\hline
\end{tabular}
\caption{\label{tab:res} Results on the \wq test set.}
\end{center}
\end{table}



We compare our system in terms of F1 score as computed by the official
evaluation script\footnote{Available from
  www-nlp.stanford.edu/software/sempre/} (F1 (Berant)) but also with a
slightly different F1 definition, termed F1 (Yao) which was used in
\cite{yao2014information} (the difference being the way that questions
with no answers are dealt with), and precision @ 1 (p@1) of the first
candidate entity (even when there are a set of correct answers),
comparing to recently published systems.\footnote{Results of baselines
  except \cite{bordes2014open} have been extracted from the original
  papers. For our experiments, all hyperparameters have been selected
  on the \wq validation set: $k$ was chosen among $\{64, 128, 256\}$,
  the learning rate on a log. scale between $10^{-4}$ and $10^{-1}$
  and we used at most $100$ paths in the subgraph representation.}
The upper part of Table~\ref{tab:res} indicates that our approach
outperforms \cite{yao2014information},
\cite{berant-EtAl:2013:EMNLP} and \cite{bordes2014open}, and performs
similarly as \cite{berant2014semantic}.

The lower part of Table~\ref{tab:res} compares various versions of our
model. Our default approach uses the Subgraph representation for
answers and $C_2$ as the candidate answers set.
%
Replacing $C_2$ by $C_1$ induces a large drop in performance because
many questions do not have answers thatare directly connected to their
inluded entity (not in $C_1$). However, using all 2-hops connections
as a candidate set is also detrimental, because the larger number of
candidates confuses (and slows a lot) our ranking based inference.
Our results also verify our hypothesis of Section~\ref{sec:hypo},
that a richer representation for answers (using the local subgraph)
can store more pertinent information.
%
Finally, we demonstrate that we greatly improve upon the model of
\cite{bordes2014open}, which actually corresponds to a setting with
the Path representation and $C_1$ as candidate set.

We also considered an ensemble of our approach and that of
\cite{berant2014semantic}.  As we only had access to their test
predictions we used the following combination method. Our approach
gives a score $S(q,a)$ for the answer it predicts. We chose a
threshold such that our approach predicts 50\% of the time (when
$S(q,a)$ is above its value), and the other 50\% of the time we
use the prediction of \cite{berant2014semantic} instead. 
We aimed for a 50/50 ratio because both methods perform similarly.
The ensemble improves the state-of-the-art, and indicates that
our models are significantly different in their design.

\section{Conclusion}
This paper presented an embedding model that learns to perform open QA
using training data made of questions paired with their answers and of
a KB to provide a structure among answers, and can achieve
promising performance on the competitive benchmark \wq.

\bibliography{../qawemb.bib}

\begin{thebibliography}{10}

\bibitem{berant-EtAl:2013:EMNLP}
J.~Berant, A.~Chou, R.~Frostig, and P.~Liang.
\newblock Semantic parsing on {Freebase} from question-answer pairs.
\newblock In {\em Proceedings of the 2013 Conference on Empirical Methods in
  Natural Language Processing}, October 2013.

\bibitem{berant2014semantic}
J.~Berant and P.~Liang.
\newblock Semantic parsing via paraphrasing.
\newblock In {\em Proceedings of the 52nd Annual Meeting of the ACL}, 2014.

\bibitem{bollacker2008freebase}
K.~Bollacker, C.~Evans, P.~Paritosh, T.~Sturge, and J.~Taylor.
\newblock Freebase: a collaboratively created graph database for structuring
  human knowledge.
\newblock In {\em Proceedings of the 2008 ACM SIGMOD international conference
  on Management of data}. ACM, 2008.

\bibitem{DBLP:journals/corr/BordesCW14}
A.~Bordes, S.~Chopra, and J.~Weston.
\newblock Question answering with subgraph embeddings.
\newblock {\em CoRR}, abs/1406.3676, 2014.

\bibitem{bordes2014open}
A.~Bordes, J.~Weston, and N.~Usunier.
\newblock Open question answering with weakly supervised embedding models.
\newblock In {\em Proceedings of ECML-PKDD'14}. Springer, 2014.

\bibitem{paralex}
A.~Fader, L.~Zettlemoyer, and O.~Etzioni.
\newblock Paraphrase-driven learning for open question answering.
\newblock In {\em Proceedings of the 51st Annual Meeting of the Association for
  Computational Linguistics}, pages 1608--1618, Sofia, Bulgaria, 2013.

\bibitem{fader2014open}
A.~Fader, L.~Zettlemoyer, and O.~Etzioni.
\newblock Open question answering over curated and extracted knowledge bases.
\newblock In {\em Proceedings of KDD'14}. ACM, 2014.

\bibitem{kolomiyets2011survey}
O.~Kolomiyets and M.-F. Moens.
\newblock A survey on question answering technology from an information
  retrieval perspective.
\newblock {\em Information Sciences}, 181(24):5412--5434, 2011.

\bibitem{kwiatkowski-EtAl:2013:EMNLP}
T.~Kwiatkowski, E.~Choi, Y.~Artzi, and L.~Zettlemoyer.
\newblock Scaling semantic parsers with on-the-fly ontology matching.
\newblock In {\em Proceedings of the 2013 Conference on Empirical Methods in
  Natural Language Processing}, October 2013.

\bibitem{lin2012entity}
T.~Lin, O.~Etzioni, et~al.
\newblock Entity linking at web scale.
\newblock In {\em Proceedings of the Joint Workshop on Automatic Knowledge Base
  Construction and Web-scale Knowledge Extraction}, pages 84--88. Association
  for Computational Linguistics, 2012.

\bibitem{recht2011hogwild}
B.~Recht, C.~R{\'e}, S.~J. Wright, and F.~Niu.
\newblock Hogwild!: A lock-free approach to parallelizing stochastic gradient
  descent.
\newblock In {\em Advances in Neural Information Processing Systems (NIPS
  24).}, 2011.

\bibitem{unger2012template}
C.~Unger, L.~B{\"u}hmann, J.~Lehmann, A.-C. Ngonga~Ngomo, D.~Gerber, and
  P.~Cimiano.
\newblock Template-based question answering over rdf data.
\newblock In {\em Proceedings of the 21st international conference on World
  Wide Web}, 2012.

\bibitem{wsabie}
J.~Weston, S.~Bengio, and N.~Usunier.
\newblock Large scale image annotation: learning to rank with joint word-image
  embeddings.
\newblock {\em Machine learning}, 81(1), 2010.

\bibitem{yao2014information}
X.~Yao and B.~Van~Durme.
\newblock Information extraction over structured data: Question answering with
  freebase.
\newblock In {\em Proceedings of the 52nd Annual Meeting of the ACL}, 2014.

\end{thebibliography}
\bibliographystyle{abbrv}

\end{document}